# Benchmarking the State of Networks with a Low-Cost Method Based on Reservoir Computing

Felix Simon Reimers[1], Carl-Hendrik Peters[1,2], and Stefano Nichele[1,3]

[1] Østfold University College, 1757 Halden, Norway
[2] Lucerne University of Applied Sciences and Arts, 6002 Luzern, Switzerland
[3] Oslo Metropolitan University, 0130 Oslo, Norway
stefano.nichele@hiof.no

**Abstract.** Using data from mobile network utilization in Norway, we showcase the possibility of monitoring the state of communication and mobility networks with a non-invasive, low-cost method. This method transforms the network data into a model within the framework of reservoir computing and then measures the model's performance on proxy tasks. Experimentally, we show how the performance on these proxies relates to the state of the network. A key advantage of this approach is that it uses readily available data sets and leverages the reservoir computing framework for an inexpensive and largely agnostic method.

Data from mobile network utilization is available in an anonymous, aggregated form with multiple snapshots per day. This data can be treated like a weighted network. Reservoir computing allows the use of weighted, but untrained networks as a machine learning tool. The network, initialized as a so-called echo state network (ESN), projects incoming signals into a higher dimensional space, on which a single trained layer operates. This consumes less energy than deep neural networks in which every weight of the network is trained. We use neuroscience inspired tasks and trained our ESN model to solve them. We then show how the performance depends on certain network configurations and also how it visibly decreases when perturbing the network. While this work serves as proof of concept, we believe it can be elevated to be used for near-real-time monitoring as well as the identification of possible weak spots of both mobile communication networks as well as transportation networks.

# 1   Introduction

Spotting faults and bottlenecks in communication networks or power grids and managing transportation networks are of high interest and increasingly data-driven, with a growing amount of machine learning related approaches being used in those fields [1,2,3]. This paper presents a proof-of-concept for an approach that relies on a computational perspective to infer knowledge about the given network - in our case the transfer rate between municipalities in Norway as measured by the cell tower network.

On the technical side the approach relies on the framework of reservoir computing, or more explicitly echo state networks [4]. Conceptually, we take a computational perspective, treating the network data as a computational substrate. While there is extensive work on fault detection in communication and other networks as well as on the identification of bottlenecks, there is to our knowledge no approach using the reservoir computing framework as we do. Instead of using graph theory to examine the networks itself or using rule-based or data-driven approaches that aim at classifying the state of the network, we aim at embedding the network in a model with as little changes as possible and then using the performance of the model as a benchmark to infer the state of the network itself. As such our overarching research question is: How does the computational performance of a given network reflect the state of that network? The content of the tasks itself is not important, as it is just a proxy. Important is that the performance of the model on those tasks can be used as a benchmark.

The approach is easily adaptable to different networks, as long as data in the form of an adjacency list or matrices is present. It is agnostic to the specific hardware and software technologies used in the network. Furthermore, the resulting models can be trained with little effort compared to artificial neural networks.

While our work is on a more conceptual level, we believe it to be a possible step towards a framework that helps to examine different networks of civil infrastructure, to assess the state they are in and to improve robustness and throughput. In order to test this, our framework needs to be integrated with further data like network failure logs to test the model.

The code to reproduce our results can be found in the following anonymized repository: https://github.com/Carasar/nor_conn2res_public.

The rest of the paper is structured as follows: In the second section, reservoir computing in general and the conn2res toolbox in particular are explained and some related work is presented. The third section explains our framework in more detail, while the fourth section explains our experimental setup. In the fifth section, the results of our work are presented and in the sixth section the results and possible future work are discussed.

## 2 Background and Related Work

This chapter aims at giving the reader an overview of reservoir computing, the used toolbox and lastly some selected related works.

### 2.1 Reservoir Computing

The concept of reservoir computing was developed independently in [4,5], naming the resulting models liquid state machine and echo state network respectively. We are using the second term throughout this paper, as this model is based on recurrent neural networks with weight-matrix multiplications and activation functions, while the liquid state machine has the - neurologically more plausible - integrate-and-fire neurons. As is pointed out in [5] and has been surveyed in [6], reservoir computing is largely agnostic to the used reservoir, as long as its internal dynamics follow certain rules, which means that a literal bucket of water or other physical systems can be used as reservoirs. When using a recurrent network as we do, the principle is as follows: A recurrent network is initialized, and its topology and weights are frozen. Input data, often in the form of timeseries, is multiplied with input weights and then passed to the nodes in the network iteratively in every time step. As the signal traverses through the recurrent network with every time step, it is being multiplied with the weights of the network and integrated and evaluated by the activation function at every node. Eventually, nodes are read out by an additional layer of neurons. Only the weights of this read-out layer are trained for the given task [4]. As only relatively few nodes have to be trained, the computational effort for training the model is much smaller than for a standard artificial neural network. A visualization of such an echo state network can be seen in figure 1.

A specific hyperparameter of an echo state network is the alpha value. This value scales all the weights within the reservoir. A specific choice of an alpha value can, if other criteria are met, put the network into the name giving echo state, which grants the best computational performance. For mathematical details, the reader is referred to [4].

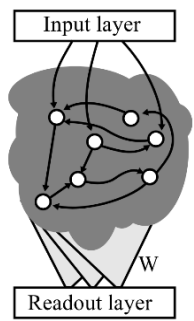

**Fig. 1.** A visualization of the reservoir computing model. Taken with approval from [7].

## 2.2 Conn2Res and Benchmarks

As previously discussed, reservoir computing is rather agnostic with regards to the used reservoir. The conn2res toolbox [8] has been designed to allow an easy implementation of connectomes as echo state networks or liquid state machines. A connectome is a mapping of the connections within biological neurons. For this reason, the toolbox allows to choose specific input and output nodes for the reservoir based on a labeling, which could for example be different regions of the brain. While our network is not biological, conn2res can be used without impediments for our purposes.

The toolbox also includes are number of implemented tasks that the model can directly be trained on. While one of them has been originally implemented, the majority stem from the reservoirpy [9] and the neurogym [10] repositories. Especially the neurogym tasks take inspiration from neuroscience. The two tasks chosen by us, the Perceptional Decision Making and the Go/No-Go task are both from the latter repository.

Our choice of neuro-inspired benchmarks is motivated by our ongoing research on the performance of biological or biology-inspired computational substrates. The used benchmarks are testing the ability of the model to measure a signal over time, integrate the signal and make a decision based on this. As such, we believe that neuro-inspired tasks can be used to benchmark a broad spectrum of potential computational abilities of reservoir computation models.

## 2.3 Related Works

The paper [6] grants an overview of different models that can be used within the reservoir computing framework. It also takes a computational perspective on the matter akin to ours.

Fault management in networks with machine learning based approaches are reviewed in [11], where fault management is described as an expensive task. AI support started with expert systems in the 1990s; more recent approaches based on data science are divided by the authors into data mining and machine learning. The described data-mining approaches include techniques resulting in graphs [12], but those are the product of their examination, whereas in our approach an existing weighted graph is used as the starting point of the model. Data mining approaches are focused on finding patterns, whereas we assume that characteristics of the network to be implicitly reflected in the computational performance. Following the definitions in [13], our approach might be used in the fault detection or localization of running systems. Furthermore, the measured degradation might help to estimate the severity of a fault or in the identification of bottlenecks. Similar to [14], our approach does not rely on a priori knowledge of the expected faults.

Multipath transport problems have to handle the existence of bottlenecks. [15] presents the first learning-based algorithm to tackle this task. While our framework does not solve pathing problems, it may help at identifying bottlenecks. This is also a relevant problem when dealing with power grids [2].

Previous work [16] has used telecommunications data to explore traffic patterns. As they describe, mobile network data is a rich source of information. While their work focuses on a smaller scale, both temporally and spatially, our aggregated data set covers the whole of Norway at the commune level over a quarter of a year. As our framework is agnostic to the network used, we believe it could be applied to such smaller scales too.

## 3   Framework

In this section we will outline how we implemented an echo state network based on our network data, which will be explained further. Next, the used benchmarks are explained in more detail.

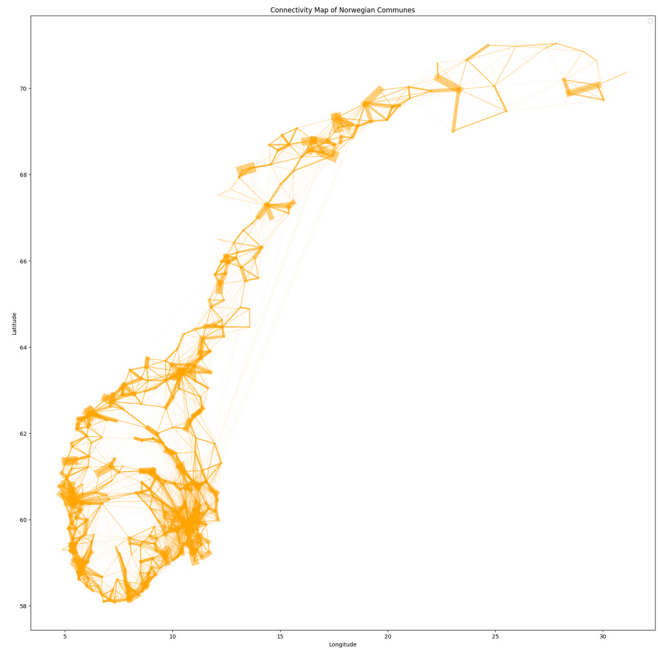

**Fig. 2.** A visualization of the mobile network utilization data. The individual nodes are arranged according to their longitude/latitude coordinates. The shape of mainland Norway is clearly visible, as is increased fluctuation around Oslo and the western coast.

### 3.1   Echo State Network

As introduced above, the reservoir computing framework allows to implement a given network as an echo state network with relative ease. Transportation networks, power grids, cell tower networks and others can be described mathematically as graphs consisting of edges and nodes. Both nodes and edges can be assigned further attributes. In our case, the transfer rates are treated as the weights of the edges, similar to the weights in an artificial neural network. Other possible values that could be assigned are for example the distances between cell towers. The nodes in our model integrate the incoming weighted signals and apply an activation function. As the transfer rates are measured from source to target municipality (instead of just measuring the net transfer), our graph contains directed edges, but there can be edges going in opposing directions with different weights.

### 3.2 Network Data

The network data we use consists of adjacency lists between the 356 communes in Norway. For two days, Tuesday and Saturday, the transfer rate measured between the cell towers of a communes is averaged for the first quarter of 2021. As a measurement period goes for six hours, four datasets for every day exist, resulting in eight adjacency lists in total. Additionally, the longitude and latitude of the communes can be used for visualizations, which allows to visualize the network over a map of Norway, as done in figure 2.

### 3.3 Neuro-inspired tasks

The tasks used in our approach are taken from the neurogym [10] repository and are inspired by the field of neuroscience. The data sets can be generated and are used as a benchmark of the computational performance of our model. The tasks can be clustered as classification, regression and memorization. The two tasks used, Perceptual Decision Making and Go/No-go are both classification tasks. They consist of time series which the model has to integrate or remember in order to score well.

In Perceptual Decision Making, the input data consists of multiple time series which can be split into segments, as can be seen in figure 3. Within every segment, the signals $x_2$ and $x_3$ are noisy timeseries of which one is one average larger than the other. $x_1$ is the fixation whose value indicates whether or not a decision has to be made in this timestep. $y$ is the ground truth: It is zero most of the time, but as a segment comes to its end and $x_1$ drops to zero, $y$ either becomes "1" or "2", depending on whether the signal $x_2$ or $x_3$ is larger on average in this segment. To perform well in this task, a model has to integrate the two signals over the timesteps of a segment, and as the fixation drops has to output the right class.

The Go/No-go task equally consists of three input signals, a fixation $x_1$ and the two timeseries $x_2$ and $x_3$ and a ground truth $y$. It is again split into segments. At the start of a segment, either $x_2$ or $x_3$ spike, giving a "go" or "no-go" signal. Some timesteps later, the model has to classify which signal was perceived earlier.

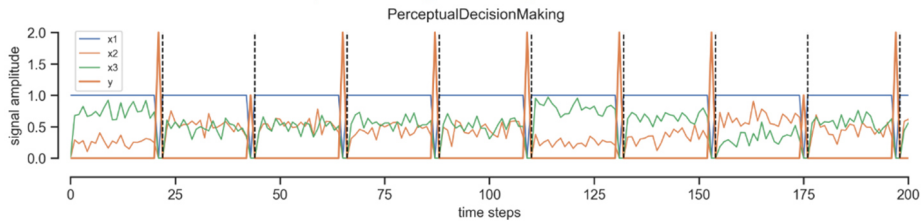

**Fig. 3.** The timeseries of the input data ($x_1$, $x_2$, $x_3$) and output data ($y$) of the Perceptual Decision Making task. As can be seen, the time series can be split into segments and at the end of each segment a classification for class "1" or "2" has to be made.

## 4 Experimental Details

In our model, the input signal is not passed to all nodes. Instead, a number of nodes corresponding to the dimension of the input, i.e. three nodes, are chosen for input and a total of six nodes are connected from the reservoir to the read-out layer. The choice of the nodes is either random, or informed by the in- and out-degree, where the nodes with the highest out-degree are chosen as input nodes and the one with the highest in-degree as output nodes. The former corresponds to a situation without additional information about the network. In the latter case, some knowledge about the network is assumed, allowing the selection of nodes that maximize or increase signal transmission through the network. This in turn should increase the performance of the model. Signals passed to or from the reservoir are not multiplied with a weight. For the activation function inside we chose $tanh$.

As mentioned above, the reservoir's weights can be scaled with an alpha hyperparameter. In preliminary experiments, the performance was calculated for a range of alpha values. For the results, the alpha corresponding to the best performance was chosen.

In order to contextualize the performance of a model, an appropriate baseline has to be found. For that, the connections of the original network are randomized, maintaining the degree distribution of the nodes and to some extent the weight distribution. The resulting null model has a different topology which may result in different performances. As this was not the focus of our work, we do not report on the performance of the null models, but the experimental results can be reproduced with the github repository.

We measured the performance of the model in the face of perturbations to the network. To that end, three randomly sampled nodes were deleted from the network iteratively. It was made sure that the deleted nodes are not part of the input or output nodes. In every iteration, the model was trained on task data and the performance measured.

Three loss functions were used to measure the performance: A standard f1-score and a balanced accuracy score taking the whole time series into account, including those parts where the fixation is one. The filtered accuracy score is custom and only considers those parts of the time series where the fixation is zero.

## 5 Results

In this section we compare the performance of the model with input/output nodes being chosen randomly or informed by the node degree. The model is then perturbed as described in the previous section. We start with the Go/No-go task and continue with the Perceptional Decision Task. We chose to display the results for the former task for the Saturday data set and for the latter task for the Tuesday data set. More results can be found with the github repository and qualitatively align with the presented work here.

As can be seen in figure 4 and 5, the performance of the model derived from the Saturday data on the Go/No-go task is initially high, corresponding to an unperturbed network. Iteratively perturbing the network leads to a drop in the measured performance. This is most prominent for the balanced accuracy store and least for the filtered accuracy score. In the case of the informed choice of input and output nodes, the networks from timeframe 1 and 3 on Saturday have a very visible drop in performance around step 25 with little variation as seen in figure 4, whereas the two remaining timeframes show a clear drop in average performance from time step 40 for the balanced accuracy and f1 score. In the networks with randomly chosen input and output nodes, the decrease in performance is overall less distinctive. Here, the perturbations often lead to a decrease of the average performance, but also to an increase in fluctuation such that high performance might still be reached as can be seen in figure 5. Only for the third time frame of Saturday a clear drop in performance can be observed very early around the $5^{th}$ time step. For the first and third time frame, the average performance gradually decreases with increasing variation. In the second time frame, the balanced accuracy and f1 score drop early, while the filtered accuracy oscillates.

For the Perceptual Decision Making task, the results have significantly less variation overall, with little difference between the informed and randomized choice of nodes. Again, the balanced accuracy score has the most distinct behavior, but for all timeframes except the first the other two scores drop to zero as well. Figure 6 depicts the performance for Tuesday with an informed choice of nodes. The drop in performance is visible either around the $25^{th}$ time step for timeframes two and four, for the $40^{th}$ time step for time frame one or around time step 60 for the third time frame. When the input/output nodes are chosen randomly like in figure 7, the drop occurs around the $20^{th}$ iteration for time frames one and two or in the $40^{th}$ or $45^{th}$ iteration for time frames four and three respectively.

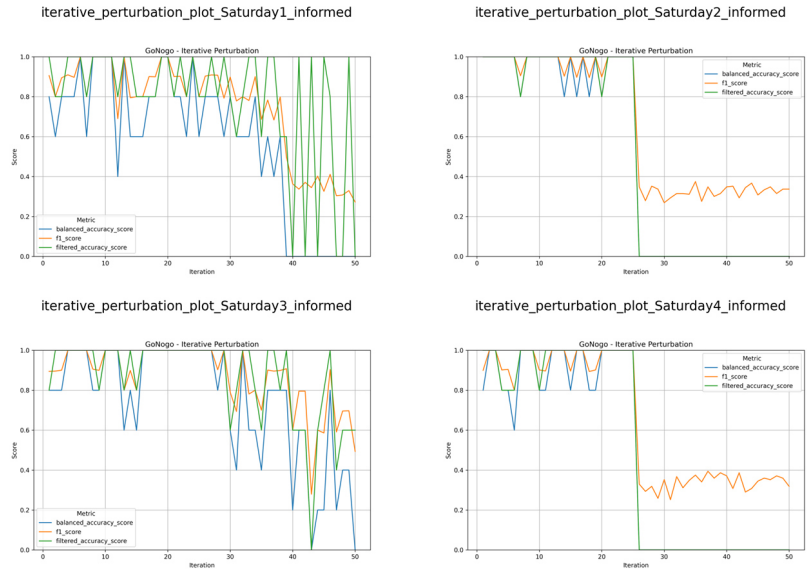

**Fig. 4.** Performance measured with balanced accuracy score, f1-score and a filtered accuracy score on the networks derived from the 4 timepoints on Saturday for the Go/No-go task. Iteratively, the network is perturbed in every step. The input and output nodes are chosen informed by the degree of the nodes.

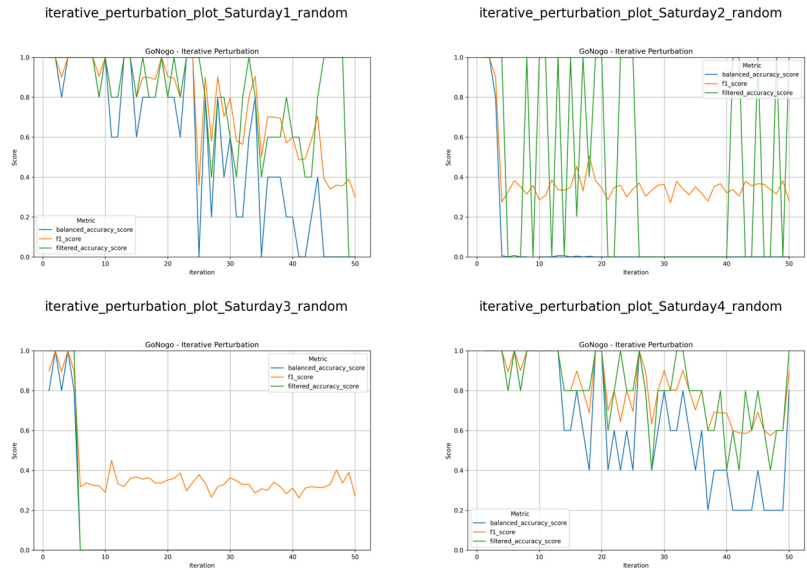

**Fig. 5.** Performance measured with balanced accuracy score, f1-score and a filtered accuracy score on the networks derived from the 4 timepoints on Saturday for the Go/No-go task. Iteratively, the network is perturbed in every step. The input and output nodes are chosen randomly.

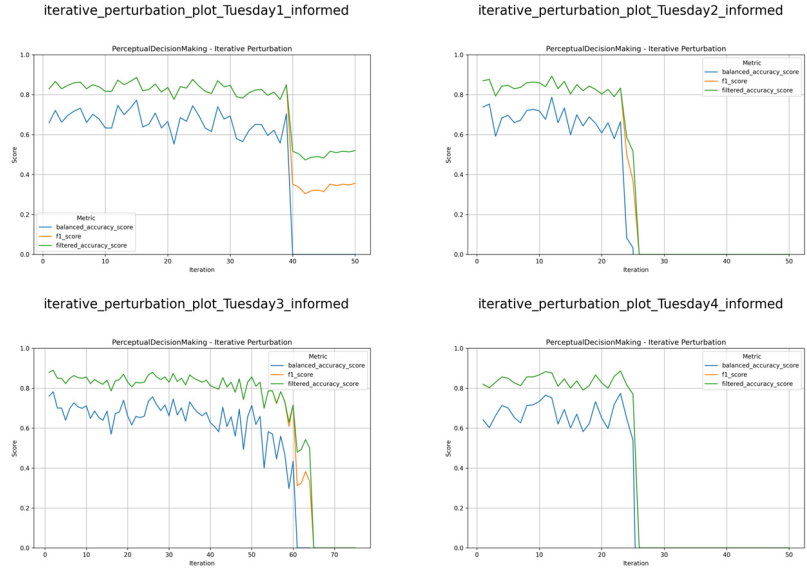

**Fig. 6.** Performance measured with balanced accuracy score, f1-score and a filtered accuracy score on the networks derived from the 4 timepoints on Tuesday for the Perceptional Decision Making task. Iteratively, the network is perturbed in every step. The input and output nodes are chosen informed by the degree of the nodes.

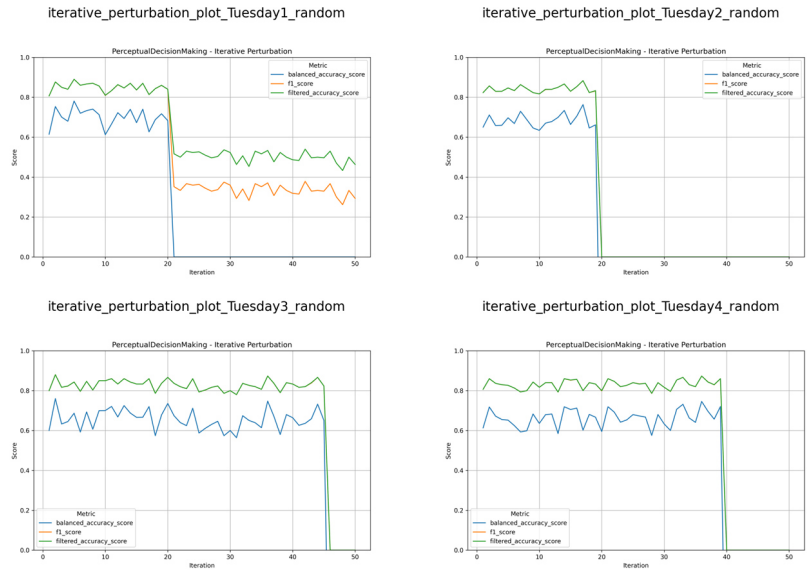

**Fig. 7.** Performance measured with balanced accuracy score, f1-score and a filtered accuracy score on the networks derived from the 4 timepoints on Tuesday for the Perceptional Decision Making task. Iteratively, the network is perturbed in every step. The input and output nodes are chosen randomly.

## 6     Discussion and Summary

The results show that the performance of our models reflects the increasing perturbation over time. The time step from which this becomes visible as well as the variation in the performance measure, depends on the chosen task as well as the time frame and the performance function. While this justifies our computational perspective, further investigation is necessary to both broaden and consolidate our results. Due to the inherent stochasticity of our framework and the difficulty of analyzing how a signal traverses a network over time, we can so far only present empirical results in this work.

By integrating more detailed network data, like failure logs, the framework can be tested in real-world settings. Application to a real-world situation will be necessary to validate the currently only conceptual work and should be a focus for future work. Furthermore, this will allow to compare our approach to other applied methods from the literature, where a focus should lie on both the efficiency but also the effectiveness of our model.

Further work should focus on examining the relationship between task performance and network state further, especially with regard to the localization of points of interest within the network. A possibility for this could be to vary the chosen nodes for input, output or artificial perturbation.

Also, the null model's performance should be included more thoroughly into the examinations, as a common baseline makes comparisons between networks of different sizes possible.

Combining the data of different networks like in [16] might allow to infer knowledge of one network through the performance benchmark of another. Trying the approach with different networks, either the same type of network form different sources or different types, like transportation networks or power grids, will show how agnostic the framework really is.

A reversed approach where nodes are added to the network to increase or stabilize the performance might be approaches that can be adopted to increase the robustness and resilience in actual networks. For example, the maximal largest power supply of a power grid after a failure might be increased [2].

## 7     Acknowledgements

We would like to thank Kenth Engø-Monsen for supplying and explaining the network utilization data set to us.

# References

1. Boutaba, Raouf, et al. "A comprehensive survey on machine learning for networking: evolution, applications and research opportunities." *Journal of Internet Services and Applications* 9.1: 1-99 (2018).
2. Tu, Haicheng, Xi Zhang, Yongxiang Xia, Fengqiang Gu, und Sheng Xu. „Bottlenecks Identification and Resilience Improvement of Power Networks in Extreme Events". *Frontiers in Physics* 10 (2022).
3. Patil, Priyadarshan. "Applications of deep learning in traffic management: A review." *International Journal of Business Intelligence and Big Data Analytics* 5.1: 16-23 (2022).
4. Jaeger, Herbert. „The "echo state" approach to analysing and training recurrent neural networks-with an erratum note'". *Bonn, Germany: German National Research Center for Information Technology GMD Technical Report* 148 (2001).
5. Maass, Wolfgang, Thomas Natschläger, and Henry Markram. "Real-time computing without stable states: A new framework for neural computation based on perturbations." *Neural computation* 14.11: 2531-2560 (2002).
6. Tanaka, Gouhei, et al. "Recent advances in physical reservoir computing: A review." *Neural Networks* 115: 100-123 (2019).
7. Nichele, Stefano, and Andreas Molund. "Deep learning with cellular automaton-based reservoir computing." (2017).
8. Suárez, Laura E., et al. "Connectome-based reservoir computing with the conn2res toolbox." *Nature Communications* 15.1: 656 (2024).
9. Trouvain, Nathan, et al. "Reservoirpy: an efficient and user-friendly library to design echo state networks." *International Conference on Artificial Neural Networks*. Cham: Springer International Publishing, (2020).
10. Croconi, Giulia, et al. "NeuroGym: An open resource for developing and sharing neuroscience tasks", https://neurogym.github.io/, last accessed 2025/05/04.
11. Nouioua, Mourad, et al. "A survey of machine learning for network fault management." *Machine Learning and Data Mining for Emerging Trend in Cyber Dynamics: Theories and Applications*. Cham: Springer International Publishing, 1-27 (2021).
12. Lozonavu, Mihaela, Martha Vlachou-Konchylaki, and Vincent Huang. "Relation discovery of mobile network alarms with sequential pattern mining." *2017 International Conference on Computing, Networking and Communications (ICNC)*. IEEE, (2017).
13. łgorzata Steinder, Ma, and Adarshpal S. Sethi. "A survey of fault localization techniques in computer networks." *Science of computer programming* 53.2: 165-194 (2004).
14. Kiciman, Emre, and Armando Fox. "Detecting application-level failures in component-based internet services." *IEEE transactions on neural networks* 16.5: 1027-1041 (2005).
15. Dong, Enhuan, et al. "SmartSBD: Smart shared bottleneck detection for efficient multipath congestion control over heterogeneous networks." *Computer Networks* 237: 110047 (2023).
16. Yusuf, Oluwaleke, Adil Rasheed, and Frank Lindseth. "Exploring urban mobility trends using cellular network data." *The International Conference on Net-Zero Civil Infrastructures: Innovations in Materials, Structures, and Management Practices (NTZR)*. Cham: Springer Nature Switzerland, 2024.